\documentclass[10pt,twocolumn,letterpaper]{article}

\usepackage{cvpr}
\usepackage{times}
\usepackage{epsfig}
\usepackage{graphicx}
\usepackage{amsmath}
\usepackage{amssymb}

\usepackage[british,english,american]{babel}

\usepackage{alex}
\usepackage{tikz,pgfplots}
\usepackage{cite}
\usepackage{tango}
\usepackage{tabulary,multirow,overpic,xcolor}
\usepackage[caption=false]{subfig}
\newcommand{\x}{\times}
\usepackage{pbox}
\usepackage{colortbl}

\makeatletter
\newcommand*\verybigcdot{\mathpalette\verybigcdot@{1.4}}
\newcommand*\verybigcdot@[2]{\mathbin{\vcenter{\hbox{\scalebox{#2}{$\m@th#1\bullet$}}}}}

\newcommand*\bigcdot{\mathpalette\bigcdot@{.7}}
\newcommand*\bigcdot@[2]{\mathbin{\vcenter{\hbox{\scalebox{#2}{$\m@th#1\bullet$}}}}}

\newcommand*\medcdot{\mathpalette\medcdot@{.5}}
\newcommand*\medcdot@[2]{\mathbin{\vcenter{\hbox{\scalebox{#2}{$\m@th#1\bullet$}}}}}

\makeatother

\newcommand{\mybigdots}{\tiny $\verybigcdot\verybigcdot\verybigcdot$}

\usepgfplotslibrary{fillbetween}
\usetikzlibrary{decorations.text}

\usetikzlibrary{arrows}
\usetikzlibrary{arrows.meta}
\usetikzlibrary{patterns}
\usetikzlibrary{3d}

\usetikzlibrary{fadings}
\tikzfading[name=myfading, bottom color=transparent!100, top color=transparent!50]

\pgfplotsset{compat = 1.3,
          legend style={font=\scriptsize},
          legend cell align={left},
          legend style={cells={align=left}, draw=black!20},
          tick style={draw=none},
          enlarge x limits=false,
          enlarge y limits=false,
          axis line style={draw=black!100},
          }

\pgfplotscreateplotcyclelist{mystylelist}{%
densely dashdotted\\
solid\\
dotted\\
densely dashed\\
dashed\\
dashdotted\\
densely dotted\\
densely dash dot dot\\
dotted\\
dash dot dot\\
}
\pgfplotscreateplotcyclelist{mystylelistv2}{%
densely dashdotted\\
dashed\\
dotted\\
densely dashed\\
solid\\
dashdotted\\
densely dotted\\
densely dash dot dot\\
dotted\\
dash dot dot\\
}
\pgfplotscreateplotcyclelist{mycolorlist}{%
chameleon3\\
skyblue2\\
orange2\\
scarletred3\\
plum1\\
aluminium6\\
skyblue3\\
chocolate2\\
maroon\\
butter3\\
}
\pgfplotscreateplotcyclelist{mycolormarklist}{%
chameleon3,mark=diamond*,mark options={solid, fill=chameleon3}\\
skyblue2,mark=pentagon*,mark options={solid}\\
scarletred3,mark=triangle*,mark options={solid}\\
plum1,mark=square*,mark options={solid}\\
aluminium5,mark=otimes*,mark options={solid}\\
orange1,mark=square*,mark options={solid}\\
skyblue3\\
chocolate2\\
maroon\\
butter3\\
}
\pgfplotscreateplotcyclelist{mycolormarklistv2}{%
chameleon3,mark=diamond*,mark options={solid, fill=chameleon3}\\
aluminium5,mark=otimes*,mark options={solid}\\
scarletred3,mark=triangle*,mark options={solid}\\
plum1,mark=square*,mark options={solid}\\
skyblue2,mark=pentagon*,mark options={solid}\\
orange1,mark=square*,mark options={solid}\\
skyblue3\\
chocolate2\\
maroon\\
butter3\\
}

\newcolumntype{x}[1]{>{\centering\arraybackslash}p{#1pt}}

\newlength\savewidth\newcommand\shline{\noalign{\global\savewidth\arrayrulewidth
  \global\arrayrulewidth 1pt}\hline\noalign{\global\arrayrulewidth\savewidth}}
\newcommand{\tablestyle}[2]{\setlength{\tabcolsep}{#1}\renewcommand{\arraystretch}{#2}\centering\footnotesize}
\makeatletter\renewcommand\paragraph{\@startsection{paragraph}{4}{\z@}
  {.5em \@plus1ex \@minus.2ex}{-.5em}{\normalfont\normalsize\bfseries}}\makeatother
\definecolor{demphcolor}{RGB}{100,100,100}
\newcommand{\demph}[1]{\textcolor{demphcolor}{#1}}
\newcommand{\app}{\raise.17ex\hbox{$\scriptstyle\sim$}}
\newcommand{\mypm}[1]{{\tiny{{\demph{{$\pm$#1}}}}}}

\usepackage[pagebackref=true,breaklinks=true,letterpaper=true,colorlinks,bookmarks=false]{hyperref}

\cvprfinalcopy %

\def\cvprPaperID{****} %

\begin{document}

\title{A Multigrid Method for Efficiently Training Video Models}

\author{%
Chao-Yuan Wu$^{1,2}$ \quad\quad\quad Ross Girshick$^2$ \quad\quad\quad Kaiming He$^2$ \\
\ \ \quad Christoph Feichtenhofer$^2$\ \quad\quad\quad\quad Philipp Kr\"ahenb\"uhl$^1$\quad\  \vspace{.5em} \\
$^1$The University of Texas at Austin  \quad $^2$Facebook AI Research (FAIR) \vspace{-.5em}
}

\maketitle
\begin{abstract}
\vspace{-2.5mm}
Training competitive deep video models is an order of magnitude slower than training their counterpart image models.
Slow training causes long research cycles, which hinders progress in video understanding research.
Following standard practice for training image models, video model training has used a fixed mini-batch shape: a specific number of clips, frames, and spatial size.
However, what is the optimal shape?
High resolution models perform well, but train slowly.
Low resolution models train faster, but are less accurate.
Inspired by \mbox{multigrid} methods in numerical optimization, we propose to use variable mini-batch shapes with different spatial-temporal resolutions that are varied according to a schedule.  The different shapes arise from resampling the training data on multiple sampling grids. Training is accelerated by scaling up the mini-batch size and learning rate when shrinking the other dimensions.
We empirically demonstrate a general and robust grid schedule that yields a significant out-of-the-box training speedup without a loss in accuracy for different models (I3D, non-local, SlowFast), datasets (Kinetics, Something-Something, Charades), and training settings (with and without pre-training, 128 GPUs or 1 GPU).
As an illustrative example, the proposed multigrid method trains a ResNet-50 SlowFast network 4.5$\x$ faster (wall-clock time, same hardware) while also improving accuracy (+0.8\% absolute) on Kinetics-400 compared to baseline training.  Code is available online.\footnote{\href{https://github.com/facebookresearch/SlowFast/blob/master/projects/multigrid}{github.com/facebookresearch/SlowFast/blob/master/projects/multigrid}}
\end{abstract}

\vspace{-2.5mm}
\section{Introduction}

Training deep networks (CNNs~\cite{lecun1989backpropagation}) on video is more computationally intensive than training 2D CNN image models, potentially by an order of magnitude.
Long training time slows progress in video understanding research, hinders scaling out to real-world data sources, and consumes significant amounts of energy and hardware.
Is this slow training unavoidable, or might there be video-specific optimization strategies that can accelerate training?

3D CNN video models are trained using mini-batch optimization methods (\eg, SGD) that process one mini-batch per iteration.
The mini-batch shape $B{\x}T{\x}H{\x}W$\footnote{We omit the channel dimension (3 for RGB) for clarity.} (mini-batch size $\x$ number of frames $\x$ height $\x$ width) is typically \emph{constant} throughout training.
A variety of considerations go into selecting this input shape, but a common heuristic is to make the $T{\x}H{\x}W$ dimensions large in order to improve accuracy, \eg, as observed in~\cite{tran2018closer,wang2018non,slowfast}.

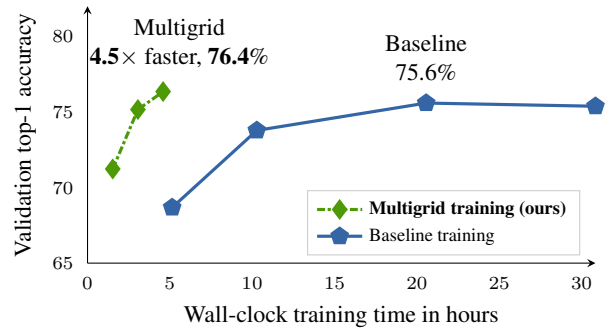
\begin{figure}[t]
\small
\begin{tikzpicture}
\begin{axis}[%
height=5cm,
width=\linewidth,
xmin=0,
ymin=65,
ymax=82,
  xlabel={Wall-clock training time in hours},
ylabel={Validation top-1 accuracy},
axis lines=left,
legend pos=south east,
nodes near coords align=north,
cycle multiindex* list={%
  mycolormarklist
      \nextlist
  mystylelist
},
every axis plot/.append style={very thick,mark options={scale=0.5, solid}},
mark size=3pt,
]
\pgfplotsset{every tick label/.append style={font=\scriptsize}}

\addplot%
table {
x y name
1.53  71.23 {0.5$\x$}
3.06  75.16 {1.0$\x$}
4.59  76.35 {1.5$\x$}
    };

\addplot%
table {
x y name
5.14  68.68 {0.25$\x$}
10.28 73.78 {0.5$\x$}
20.5644 75.5900 {1.0$\x$}
30.8467 75.3800 {1.5$\x$}
    };

\node[align=center] at (axis cs: 20.5644,78.5900) {Baseline\\75.6\%};
\node[align=center] at (axis cs: 5.59,79.5900) {Multigrid\\\textbf{4.5}$\x$ faster, \textbf{76.4}\%};

\legend{\textbf{Multigrid training (ours)}, Baseline training}
\end{axis}
\end{tikzpicture}
  \vspace{2mm}
  \caption{\textbf{Training time \vs top-1 accuracy on Kinetics-400} with a ResNet-50 SlowFast network.
  Each point corresponds to a model trained for a specific number of epochs.
  \textbf{Multigrid training}, the method developed in this paper, obtains a significantly better trade-off than baseline training.
  For example, under default settings, multigrid training is \textbf{4.5}$\x$ faster while achieving higher (+0.8\% absolute) top-1 accuracy. \emph{All methods here, and throughout the paper, use the same hardware and software implementation.}
  }\label{fig:teaser}
\end{figure}

This heuristic is only one possible choice, however, and in general there are trade-offs.
For example, one may use a smaller number of frames and/or spatial size while simultaneously increasing the mini-batch size $B$.
With such an exchange, it is possible to process the same number of epochs (passes over the dataset) with lower wall-clock time because each iteration processes more examples.
The resulting trade-off is faster training with lower accuracy.

The central idea of this paper is to avoid this trade-off---\ie, to have faster training \emph{\mbox{without}} losing accuracy---by making the mini-batch shape \emph{\mbox{variable}} during training.
By viewing the input video clips in a mini-batch as raw video signals that are sampled on a sampling grid (to be defined), we can draw a connection to \emph{\mbox{multigrid methods}} for numerical analysis~\cite{briggs2000multigrid}.
These methods exploit coarse-to-fine grids to accelerate optimization.
Intuitively, if we use large mini-batches with relatively small time and space dimensions (a `coarse grid') early in training and small mini-batches with large time and space dimensions (a `fine grid') later, then SGD may be able to scan through the data more quickly on average while finally solving for a high accuracy model, akin to how coarse grids enable solving problems on finer grids more rapidly in multigrid numerical solvers \cite{briggs2000multigrid}.

Multigrid training is possible because video models are compatible with input data of variable space and time dimensions due to weight sharing operations (\eg, convolutions).
In addition, CNNs are effective at learning patterns at multiple scales, \eg, as observed when training with data augmentation~\cite{krizhevsky2012imagenet,simonyan2014very,resnet}.
We observe similar multi-scale robustness and generalization with multigrid training.

Our proposed multigrid training method is simple and effective.
It is easy to implement and typically only requires small changes to a data loader.
Empirically, it works with default learning rate schedules and hyper-parameters already in use. No tuning is required.
Moreover, multigrid training works robustly out-of-the-box for different models (I3D~\cite{carreira2017quo}, non-local~\cite{wang2018non}, SlowFast~\cite{slowfast}), datasets (Kinetics-400~\cite{kay2017kinetics}, Something-Something V2~\cite{ssv2}, and Charades~\cite{charades}), initializations (random and pre-trained), and hardware scales (\eg, 128 GPUs or 1 GPU).
We observe a consistent speedup and performance gain in all cases without tuning.
As an example, we train a SlowFast network $\app$4.5$\x$ faster in wall-clock time on the large-scale Kinetics dataset (\figref{teaser}) while also reaching a higher accuracy (+0.8\% absolute).
We hope these benefits provided by multigrid training will make research on video understanding more \emph{accessible, scalable, and economical}.

\section{Related Work}

\paragraph{3D CNN video models}
extend 2D CNNs to model both spatial and temporal patterns.
They are currently the state of the art for video understanding~\cite{tran2015learning,carreira2017quo,xie2018rethinking,qiu2017learning,wang2018non,tran2018closer,girdhar2019video,wu2019long,tran2019video,piergiovanni2019evolving,hussein2019timeception,luo2019grouped,slowfast,martinez2019action}.
These methods are computationally expensive, both for training and inference~\cite{tran2015learning,xie2018rethinking}.
Some recent studies propose lighter weight models that use
efficient temporal modules~\cite{qiu2017learning,tran2019video,lin2018temporal,chen2018multi,lee2018motion,carreira2018massively,hussein2019timeception,piergiovanni2019evolving,sun2015human,wang2018appearance,luo2019grouped,jiang2019stm} and/or exploit temporal redundancy~\cite{zolfaghari2018eco,slowfast}.
In this paper, we show that the training time of state-of-the-art efficient models~\cite{slowfast} can still be reduced significantly.

\paragraph{Efficient training}
can also be advanced through, \eg, optimization methods (\eg,~\cite{duchi2011adaptive,kingma2014adam,smith2018dontdecay,qiao2019neural}), pre-training~\cite{carreira2017quo,ghadiyaram2019large},
distributed training~\cite{goyal2017accurate,you2018imagenet},
or advances in hardware~\cite{jouppi2017datacenter} and software design~\cite{chetlur2014cudnn,chen2018tvm}.
In this paper, we propose a complementary direction that exploits variable mini-batch shapes for fast training. Related to our method, Wang~\etal~\cite{wang2018non} and Feichtenhofer~\etal~\cite{feichtenhofer2016spatiotemporal}
initialize larger models with smaller, fully-trained ones.
These methods can potentially speed up training as well, and (as can be seen later) are a special case of multigrid training.

\paragraph{Multi-scale training}
in segmentation~\cite{he2017mask} and classification~\cite{simonyan2014very,resnet} uses multiple image crop sizes.
However, the mini-batch shape remains fixed~\cite{he2017mask,simonyan2014very,resnet}.
Multigrid training on the other hand uses variable mini-batch shapes.
He~\etal~\cite{he2015spatial} change the input shapes, but fix the mini-batch size.
These methods shows that training with variable scales can be beneficial.
Multigrid training enjoys the same property.

\paragraph{Multigrid methods}
were originally proposed for numerical boundary value problems,
and later developed into an entire field in computational mathematics~\cite{briggs2000multigrid}.
They typically involve iterating through cycles of coarse and fine problems, and exploit the fact that a coarse problem can be solved efficiently to speed up the overall problem solving.
He and Xu~\cite{he2019mgnet} connect multigrid methods to deep networks through identifying the correspondence between steps in traditional multigrid methods and operators in a convolutional neural network.
In this paper, we take inspiration from multigrid concepts from a more abstract view to accelerate video model training.

\section{Multigrid Training for Video Models}

To develop our multigrid training method we will consider a reference video model (\eg, C3D~\cite{tran2015learning}, I3D~\cite{carreira2017quo}) that is trained by a baseline mini-batch optimizer (\eg, SGD) that operates on mini-batches of shape $B{\x}T{\x}H{\x}W$ (mini-batch size $\x$ number of frames $\x$ height $\x$ width) for some number of epochs (\eg, 100).
The spatial-temporal shape, $T{\x}H{\x}W$, arises from resampling source videos in the training dataset according to a \emph{sampling grid} that is specified by a temporal span, a spatial span, a temporal stride, and a spatial stride (defined in \secref{concepts}). These concepts intuitively correspond to a grid's duration/area (span) and sampling rate (stride).
The baseline optimizer holds the mini-batch shape \emph{constant} across all training iterations.

\paragraph{Proposed Multigrid Method.} Inspired by multigrid methods in numerical analysis, which solve optimization problems on alternating coarse and fine grids, the core observation in this paper is that the underlying sampling grid that is used to train video models need not be constant during training. In fact, we will show in experiments that by \emph{varying} the sampling grid \emph{and} the mini-batch size during training it is possible to reduce training complexity substantially (in terms of total FLOPs and wall-clock time) while achieving similar accuracy in comparison with the baseline.

The fundamental concept that enables multigrid training is the balance between computation allocated to processing more examples per mini-batch \vs the computation allocated to processing larger time and space dimensions. To control this balance, we will consider temporal and spatial shapes $t{\x}w{\x}h$ that are formed by resampling source videos with a new sampling grid that has its own spans and strides. When changing the input shape we use a scaled mini-batch size $b$ satisfying the relation $b{\cdot}t{\cdot}h{\cdot}w = B{\cdot}T{\cdot}H{\cdot}W$, or
\begin{align}
b = B \frac{T}{t} \frac{H}{h} \frac{W}{w},\label{eq:bsscale}
\end{align}
which yields computation (in FLOPs) that is roughly equal to the computation of the aforementioned baseline mini-batch for typical 3D CNNs.\footnote{
  In practice, the computation is not exactly equal, because of rounding (\eg, $w$ can be $\frac{W}{\sqrt{2}}$), padding, and, \eg, fully connected or non-local layers. We ignore these subtleties and only use approximate FLOPs as a rough design principle. All speedups are measured by wall-clock time.}

Our multigrid method uses a \emph{set of sampling grids} and a \emph{grid schedule} that determines which grid to use in each training iteration.
If training is run for a similar number of epochs regardless of the choice of grids,\footnote{In practice, a similar number of epochs (\eg, within a factor of 2) are typically used for a given dataset, \emph{even for very different models}.} then by making $b{>}B$ on average the entire training process can use fewer total FLOPs and have a lower wall-clock time.

We will experimentally investigate two questions:
(i) is there a set of grids with a grid schedule that can lead to faster training without a loss in accuracy? and, (ii) if so, does it robustly generalize to new models and datasets without modification?
In the following we will develop the core multigrid training concepts in detail (\secref{concepts}), provide an implementation (\ie, a set of grids and a grid schedule) that work well in practice (\secref{implementation}), and then explore ablation and generalization experiments (\secref{kinetics}).

\subsection{Multigrid Training Concepts}
\label{sec:concepts}

\tikzset{
    conceptstyle/.style = {chameleon3,mark=*},
}

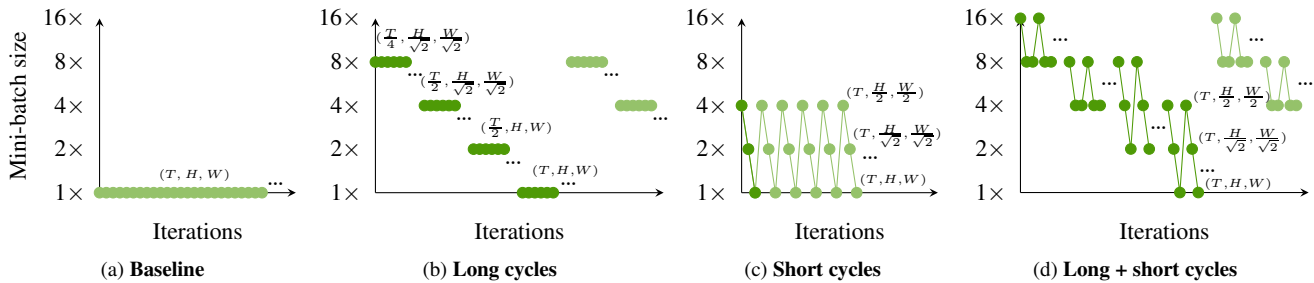
\begin{figure*}
  \small
  \subfloat[\textbf{Baseline}\label{tab:concept:baseline}]{%
  \begin{tikzpicture}
  \begin{axis}[
    clip=false,
    width=0.24\textwidth,
    height=3.9cm,
    xmin=1,
    xmax=30,
    ymin=1,
    ymax=5,
    xticklabels={,,},
    ytick={1, 2, 3, 4, 5},
    yticklabels={1$\x$,2$\x$,4$\x$,8$\x$,16$\x$},
    axis lines=left,
    xlabel={Iterations},
    ylabel={Mini-batch size},
  ]
  \addplot[chameleon3!60,mark=*]
      table[x index=0, y index=1] {plotdata/concept_figs/baseline.data};

  \node[align=center,scale=0.9] at (axis cs: 15,1.4) {\tiny $(T,H,W)$};
  \node[align=center,scale=0.3] at (axis cs: 27,1.2) {\mybigdots};
  \end{axis}
  \end{tikzpicture}}\hfill
  \subfloat[\textbf{Long cycles}\label{tab:concept:long}]{%
  \begin{tikzpicture}
  \begin{axis}[
    clip=false,
    width=0.31\textwidth,
    height=3.9cm,
    xmin=1,
    xmax=48,
    ymin=1,
    ymax=5,
    xticklabels={,,},
    ytick={1, 2, 3, 4, 5},
    yticklabels={1$\x$,2$\x$,4$\x$,8$\x$,16$\x$},
    axis lines=left,
    xlabel={Iterations},
  ]
  \addplot[conceptstyle]
      table[x index=0, y index=1] {plotdata/concept_figs/long_cycle_1.data};
  \addplot[conceptstyle]
      table[x index=0, y index=1] {plotdata/concept_figs/long_cycle_2.data};
  \addplot[conceptstyle]
      table[x index=0, y index=1] {plotdata/concept_figs/long_cycle_3.data};
  \addplot[conceptstyle]
      table[x index=0, y index=1] {plotdata/concept_figs/long_cycle_4.data};
  \addplot[chameleon3!60,mark=*]
      table[x index=0, y index=1] {plotdata/concept_figs/long_cycle_5.data};
  \addplot[chameleon3!60,mark=*]
      table[x index=0, y index=1] {plotdata/concept_figs/long_cycle_6.data};

  \node[align=center,scale=0.3] at (axis cs: 7.5,3.7) {\mybigdots};
  \node[align=center,scale=0.3] at (axis cs: 15.5,2.7) {\mybigdots};
  \node[align=center,scale=0.3] at (axis cs: 23.5,1.7) {\mybigdots};
  \node[align=center,scale=0.3] at (axis cs: 32.5,1.2) {\mybigdots};
  \node[align=center,scale=0.3] at (axis cs: 39.5,3.7) {\mybigdots};
  \node[align=center,scale=0.3] at (axis cs: 47.5,2.7) {\mybigdots};

  \node[align=center,scale=0.9] at (axis cs: 8.8,4.5) {\tiny $(\frac{T}{4}{,}\frac{H}{\sqrt{2}}{,}\frac{W}{\sqrt{2}})$};
  \node[align=center,scale=0.9] at (axis cs: 16,3.5) {\tiny $(\frac{T}{2}{,}\frac{H}{\sqrt{2}}{,}\frac{W}{\sqrt{2}})$};
  \node[align=center,scale=0.9] at (axis cs: 24,2.5) {\tiny $(\frac{T}{2}{,}H{,}W)$};
  \node[align=center,scale=0.9] at (axis cs: 32,1.5) {\tiny $(T{,}H{,}W)$};
  \end{axis}
  \end{tikzpicture}}\hfill
  \subfloat[\textbf{Short cycles}\label{tab:concept:short}]{%
  \begin{tikzpicture}
  \begin{axis}[
    clip=false,
    width=0.24\textwidth,
    height=3.9cm,
    xmin=1,
    xmax=30,
    ymin=1,
    ymax=5,
    xticklabels={,,},
    ytick={1, 2, 3, 4, 5},
    yticklabels={1$\x$,2$\x$,4$\x$,8$\x$,16$\x$},
    axis lines=left,
    xlabel={Iterations},
  ]
  \addplot[chameleon3!60,mark=*]
      table[x index=0, y index=1] {plotdata/concept_figs/short_cycle.data};

  \addplot[conceptstyle]
      table[x index=0, y index=1] {plotdata/concept_figs/short_cycle_last.data};

  \node[align=center,scale=0.3] at (axis cs: 20,1.8) {\mybigdots};

  \node[align=left,scale=0.9] at (axis cs: 22.2,3.3) {\tiny $(T{,}\frac{H}{2}{,}\frac{W}{2})$};
  \node[align=left,scale=0.9] at (axis cs: 24.0,2.3) {\tiny $(T{,}\frac{H}{\sqrt{2}}{,}\frac{W}{\sqrt{2}})$};
  \node[align=left,scale=0.9] at (axis cs: 23.4,1.3) {\tiny $(T{,}H{,}W)$};

  \end{axis}
  \end{tikzpicture}}\hfill
  \subfloat[\textbf{Long + short cycles}\label{tab:concept:longshort}]{%
  \begin{tikzpicture}

  \begin{axis}[
    clip=false,
    width=0.31\textwidth,
    height=3.9cm,
    xmin=1,
    xmax=48,
    ymin=1,
    ymax=5,
    xticklabels={,,},
    ytick={1, 2, 3, 4, 5},
    yticklabels={1$\x$,2$\x$,4$\x$,8$\x$,16$\x$},
    axis lines=left,
    xlabel={Iterations},
    set layers,%
    mark layer=axis tick labels,%
  ]
  \addplot[chameleon3,mark=*]
      table[x index=0, y index=1] {plotdata/concept_figs/long_short_cycle_1.data};
  \addplot[chameleon3,mark=*]
      table[x index=0, y index=1] {plotdata/concept_figs/long_short_cycle_2.data};
  \addplot[chameleon3,mark=*]
      table[x index=0, y index=1] {plotdata/concept_figs/long_short_cycle_3.data};
  \addplot[chameleon3,mark=*]
      table[x index=0, y index=1] {plotdata/concept_figs/long_short_cycle_4.data};
  \addplot[chameleon3!60,mark=*]
      table[x index=0, y index=1] {plotdata/concept_figs/long_short_cycle_5.data};
  \addplot[chameleon3!60,mark=*]
      table[x index=0, y index=1] {plotdata/concept_figs/long_short_cycle_6.data};
  \node[align=center,scale=0.3] at (axis cs: 7.5,4.5) {\mybigdots};
  \node[align=center,scale=0.3] at (axis cs: 15.5,3.5) {\mybigdots};
  \node[align=center,scale=0.3] at (axis cs: 23.5,2.5) {\mybigdots};
  \node[align=center,scale=0.3] at (axis cs: 31.5,1.5) {\mybigdots};
  \node[align=center,scale=0.3] at (axis cs: 39.5,4.5) {\mybigdots};
  \node[align=center,scale=0.3] at (axis cs: 47.5,3.5) {\mybigdots};

  \node[align=left,scale=0.9] at (axis cs: 35.5,3.2) {\tiny $(T{,}\frac{H}{2}{,}\frac{W}{2})$};
  \node[align=left,scale=0.9] at (axis cs: 37.2,2.2) {\tiny $(T{,}\frac{H}{\sqrt{2}}{,}\frac{W}{\sqrt{2}})$};
  \node[align=left,scale=0.9] at (axis cs: 36.5,1.2) {\tiny $(T{,}H{,}W)$};

  \end{axis}
  \end{tikzpicture}}
  \vspace{2mm}
  \caption{
    \textbf{A general and robust grid schedule (\secref{implementation}).}
   We contrast multigrid training with standard baseline training.
  (a) \textbf{Baseline} training methods typically use a fixed mini-batch shape throughout training.
  (b) \textbf{Multigrid long cycles} loop over inputs from small shapes (with large mini-batch sizes) to large shapes (with small mini-batch sizes),
  staying on each shape for several epochs.
  (c) \textbf{Multigrid short cycles} rapidly move through a variety of spatial shapes, changing at each iteration.
  (d) \textbf{Multigrid long + short cycles (our default setting)} combines long and short cycles, and moves through shapes at two frequencies simultaneously.
  {\color{chameleon3}\textbf{Dark green}} points in (b), (c), and (d) correspond to one full period of a long cycle, a full short cycle, and a long+short cycle, respectively.
}\label{fig:cycles}
\end{figure*}

\paragraph{Sampling Grids.} Each video in a dataset is a discrete signal that was sampled from an underlying continuous signal generated by the physical world. The video has some number of frames and pixels per frame, which are related to the physical world by the temporal and spatial resolution of the recording device (which depends on a number of camera properties). When using one of these source videos in a training mini-batch, a sampling grid is used to \emph{resample} it.

A sampling grid in one dimension (space or time) is defined by two quantities: a \emph{span} and a \emph{stride}. Their units are defined \wrt the source video being resampled.\footnote{Between two videos these units may have different physical meanings if the videos were captured by cameras with different properties (\eg, a 24 frame span from a 24 FPS video \vs a 24 frame span from a 30 FPS video). These properties may be unknown and therefore we define grid units with respect to source videos, not the physical world.}
For the time dimension, the units are frames while for the spatial dimensions the units are pixels. The span is the support size of the grid and defines the duration or area that the grid covers. The stride is the spacing between sampling points.
Dividing the span by the stride gives the number of points in the grid, which determines the shape of the input data. Note that different grids can yield the same data shape, which implies that the mini-batch size will only change (\eqnref{bsscale}) if a change in the sampling grid also changes the data shape.

We note that spatial sampling grids already appear in the baseline optimizer if it uses multi-scale \emph{spatial} data augmentation~\cite{simard2003best,cirecsan2011high,krizhevsky2012imagenet}. Under our multigrid perspective, multi-scale spatial data augmentation changes the spatial spans and strides of the resampling grid \emph{proportionally} so that the resulting mini-batch always has the same $H{\x}W$ spatial shape. In contrast, we will change spans and strides by different factors, which results in a different spatial shape $h{\x}w$ for each grid (and likewise for the time dimension).

\paragraph{Grid Scheduling.} We use mini-batch optimizers, which have as their most basic scheduling unit a single mini-batch iteration in which one model update is performed. The training schedule consists of some number of mini-batch iterations and is often expressed in terms of epochs. For example, training may consist of 100 or 200 epochs worth of iterations. Within this overall training schedule it is common to let the learning rate vary, such as annealing it according to a schedule defined in terms of iterations or epochs.

Scheduling other training properties is also possible. Central to our multigrid method is the idea of scheduling the sampling grids that are used throughout training. When changing grids, the mini-batch size is always scaled according to \eqnref{bsscale} so that mini-batch FLOPs are held roughly constant. Grid scheduling is highly flexible, admitting a large design space from simply cycling through a sequence of pre-defined grids to using randomized grids. In \secref{implementation} we will present a randomized, hierarchical schedule that works well in practice.

\paragraph{Multigrid Properties.} Multigrid training relies on two properties of the data and model. First, resampling the data on different grids requires a suitable operator. For video, this operator can be a reconstruction filter applied to the source discrete signal followed by computing the values at the points specified by the grid (\eg, bilinear interpolation).

Second, the model must be compatible with inputs that are resampled on different grids, and therefore might have different shapes during training. Models that are composed of functions that use weight sharing across the dimensions that are resampled, \eg, 2D and 3D convolutions, recurrent functions, and self-attention, are compatible and cover most of the commonly used architectures; fully-connected layers, unless their inputs are pooled to a fixed size, are not compatible.\footnote{If an appropriate operator exists for `resampling' model \emph{parameters} so that they are compatible with new input shapes, then these parameters may still be usable with multigrid training. This concept can be combined with weight sharing, \eg, by dilating or resizing model filters to mirror the data sampling grid, though preliminary experiments did not improve results.}
We will focus on models that use 2D and 3D convolutions, as well as self-attention operations in the form of non-local blocks~\cite{wang2018non}; all models end with global average pooling and a single full-connected layer as the classifier, as is common practice.

\paragraph{Training and Testing Distributions.} The focus of this work is on multigrid methods for \emph{training} and therefore we use a standard inference method that uses a single shape for the testing data. This choice, however, may introduce a mismatch between the data distribution used to train the model and the data distribution used at test time. To close this gap, training may be finished with some number of `fine-tuning' iterations that use grids more closely aligned with the testing distribution, \eg, see~\cite{touvron2019FixRes}. We find that this fine-tuning gives a small, but consistent improvement.

\subsection{Implementation Details}\label{sec:implementation}
Multigrid training involves a choice of sampling grids and a grid schedule, which leads to a rich design space. We use a hierarchical schedule that involves alternating between mini-batch shapes at two different frequencies: a \emph{long cycle} that moves through a set of \emph{base shapes}, generated by a variety of grids, staying on each shape for several epochs, and a \emph{short cycle} that moves through a set of shapes that are `nearby' the current base shape, staying on each one for a single iteration. This hierarchical grid schedule is described in more detail shortly and illustrated in \figref{cycles}.

The remainder of this subsection provides details for this design, which we have found to work well in practice. After presenting these details, we will explore what design decisions are important in ablation experiments.

\paragraph{Optimizer.}
We use SGD with momentum and a stepwise learning rate decay schedule since these are common choices in practice~\cite{krizhevsky2012imagenet,resnet,tran2015learning,slowfast}. Using other learning rate schedules and optimizers is also possible. Specific schedules are given in each experimental section.

\paragraph{Long Cycle.}
We use sampling grids that result in an ordered sequence of $S{=}4$ base mini-batch shapes of non-decreasing size along each dimension: $8B{\x}\frac{T}{4}{\x}\frac{H}{\sqrt{2}}{\x}\frac{W}{\sqrt{2}}$, $4B{\x}\frac{T}{2}{\x}\frac{H}{\sqrt{2}}{\x}\frac{W}{\sqrt{2}}$, $2B{\x}\frac{T}{2}{\x}H{\x}W$, and $B{\x}T{\x}H{\x}W$.
These four shapes cover an intuitive range and work well in practice.
The long cycle is synchronized with the stepwise learning rate decay schedule: a full cycle over the $S$ shapes occurs exactly once for each learning rate stage.
We train on each shape for the same number of iterations.

We use a simple randomized strategy to generate a mini-batch with the target input shape for each training iteration. For each video to be used in the mini-batch, we select a random span from a specified range and set the stride such that the desired shape is produced when sampling on the resulting grid. For the spatial dimensions, this strategy amounts to resizing a random crop to the desired shape using bilinear interpolation (similar to random cropping used in image classification~\cite{krizhevsky2012imagenet,simonyan2014very,resnet}). For the temporal dimension, this strategy amounts to selecting a random temporal crop and subsampling its frames. The sampling range for spans is specified in each experimental section.

\paragraph{Short Cycle.}
The short cycle rapidly moves through a variety of spatial shapes, changing at each iteration. By default, we use the following 3-shape short cycle. For iteration $i$, let $m = i \pmod 3$; if $m{=}0$, then we set the spatial shape to $\frac{H}{2}{\x}\frac{W}{2}$; if $m{=}1$, we use $\frac{H}{\sqrt{2}}{\x}\frac{W}{\sqrt{2}}$; otherwise, the current base spatial shape from the long cycle is used.

The short cycle can be applied on its own or in conjunction with the long cycle. The mini-batch size is again scaled using \eqnref{bsscale}. The same randomized grid strategy is applied to sample data for the target mini-batch shape.

\paragraph{Learning Rate Scaling.}
When the mini-batch size changes due to the long cycle, we apply the linear scaling rule~\cite{goyal2017accurate} to adjust the learning rate by the mini-batch size scaling factor (thus either 8$\x$, 4$\x$, 2$\x$, or 1$\x$).
We found that this adjustment is harmful if applied to mini-batch size changes due to the short cycle and therefore we only adjust the learning rate when the long cycle base shape changes.

\paragraph{Fine-tuning Phase.}
If the baseline optimizer uses $L$ learning rate (LR) stages, then we apply the long and short cycles in the first $L{-}1$ LR stages. We use the corresponding $L$-th stage for fine-tuning to help match the training and testing distributions, similar to~\cite{touvron2019FixRes}.
In the first half of the fine-tuning iterations we use the $L{-}1$-st learning rate and in the second half we use the final ($L$-th) learning rate.
While fine-tuning we use the short cycle (as data augmentation), but not the long cycle.

\paragraph{Batch Normalization.}
The behavior of Batch Normalization (BN)~\cite{ioffe2015batch} depends on mini-batch statistics.
In traditional trainers,
the constant mini-batch size is also a hyper-parameter that impacts BN behaviors (\eg, the noisiness of the statistics). As our multigrid method uses variable mini-batch sizes, it is desirable to \emph{decouple} its impact on BN from that of training speedup. The following heuristic works well in practice:
we compute BN statistics with a \mbox{standardized} \emph{sub-mini-batch} of size 8; when the short cycle increases the overall mini-batch size by 2$\x$ or 4$\x$, we likewise increase the BN sub-mini-batch size to 16 and 32, respectively.

\begin{figure*}[t]
\begin{tikzpicture}
\begin{axis}[%
grid=both,
grid style={dotted},
clip=false,
xmin=0,
ymin=65,
ymax=78.4,
width=\linewidth,
height=5.75cm,
xlabel={Wall-clock time in hours (same hardware and software implementation for all methods)},
ylabel={Validation top-1 accuracy},
axis lines=left,
legend pos=south east,
nodes near coords align=south,
cycle multiindex* list={%
  mycolormarklistv2
      \nextlist
  mystylelistv2
},
every axis plot/.append style={very thick,mark options={scale=0.5, solid}},
mark size=2.5pt,
]
\pgfplotsset{every tick label/.append style={font=\scriptsize}}

\addplot
table {
x y name
1.53  71.23 {0.5$\x$}
3.06  75.16 {1.0$\x$}
4.59  76.35 {1.5$\x$}
6.12  76.20 {2.0$\x$}
    };

\addplot
table {
x y name
2.59  72.32 {0.5$\x$}
5.18  74.96 {1.0$\x$}
7.77  75.18 {1.5$\x$}
10.36 75.21 {2.0$\x$}
    };

\addplot
table {
x y name
3.80  70.26 {0.5$\x$}
7.60  75.45 {1.0$\x$}
11.39 76.28 {1.5$\x$}
15.19 77.02 {2.0$\x$}
22.79 76.51 {3.0$\x$}
};

\addplot
table {
x y name
0.87  66.15 {0.5$\x$}
1.73  69.10 {1.0$\x$}
2.60  69.90 {1.5$\x$}
3.47  70.00 {2.0$\x$}
};

\addplot
table {
x y name
5.14  68.68 {0.25$\x$}
10.28 73.78 {0.5$\x$}
20.5644 75.5900 {1.0$\x$}
30.8467 75.3800 {1.5$\x$}
41.13 74.92 {2.0$\x$}
    };

\addplot[
only marks,
nodes near coords*={\tiny \Name},
visualization depends on={value \thisrow{name} \as \Name},
]
table {
x y name
1.53  71.23 {0.5$\x$}
3.06  75.16 {1.0$\x$}
4.59  76.35 {\underline{\textbf{1.5}$\x$}}
6.12  76.20 {2.0$\x$}
7.60  75.45 {1.0$\x$}
11.39 76.28 {1.5$\x$}
15.19 77.02 {2.0$\x$}
22.79 76.51 {3.0$\x$}
0.87  66.15 {0.5$\x$}
1.73  69.10 {1.0$\x$}
2.60  69.90 {1.5$\x$}
};

\addplot[
only marks,
nodes near coords*={\tiny \Name},
nodes near coords align=west,
visualization depends on={value \thisrow{name} \as \Name},
]
table {
x y name
3.80  70.26 {0.5$\x$}
2.59  72.32 {0.5$\x$}
10.36 75.21 {2.0$\x$}
};

\addplot[
only marks,
nodes near coords*={\tiny \Name},
nodes near coords align=north,
visualization depends on={value \thisrow{name} \as \Name},
]
table {
x y name
7.77  75.18 {1.5$\x$}
5.18  74.96 {1.0$\x$}
3.47  70.00 {2.0$\x$}
20.5644 75.5900 {\underline{\textbf{1.0}$\x$}}
30.8467 75.3800 {1.5$\x$}
41.13 74.92 {2.0$\x$}
5.14  68.68 {0.25$\x$}
};

\addplot[
only marks,
nodes near coords*={\tiny \Name},
nodes near coords align=north west,
visualization depends on={value \thisrow{name} \as \Name},
]
table {
x y name
10.28 73.78 {0.5$\x$}
};

\legend{\textbf{Multigrid} (\underline{default}: long+short cycle),\textbf{Multigrid} (long cycle only),\textbf{Multigrid} (short cycle only),Baseline (constant shape: $8B{\x}\frac{T}{4}{\x}\frac{H}{\sqrt{2}}{\x}\frac{W}{\sqrt{2}}$), Baseline (\underline{default} constant shape: $B{\x}T{\x}H{\x}W$)}

\node[align=center] at (axis cs: 20.5644,74) {\footnotesize Default baseline};
\node[align=center] at (axis cs: 4.59,78) {\footnotesize Default multigrid};
\end{axis}
\end{tikzpicture}
  \caption{\textbf{Multigrid \vs baseline training.}
  Each point corresponds to one model trained with a specific schedule choice.
  Annotations denote training epochs relative to the baseline 1.0$\x$ schedule.
  For example, `1.5$\x$' denotes training for 1.5$\x$ more epochs than the default `1.0$\x$' baseline schedule (112k iterations or $\app$239 epochs).
  We see that \textbf{all variants of multigrid training achieve a better trade-off than baseline training, which uses a constant mini-batch shape}.
  Also note that multigrid training can iterate through the same number of epochs more efficiently.
  }\label{fig:exp:main}
\end{figure*}
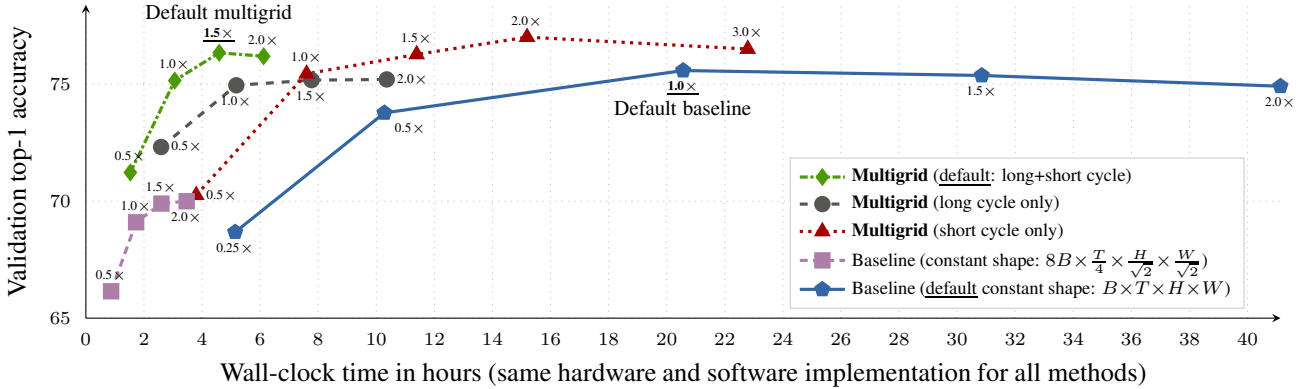

\section{Experiments on Kinetics}\label{sec:kinetics}
We conduct ablation studies on the Kinetics-400 dataset~\cite{kay2017kinetics}, which is used in prior research and requires classifying each video into one of 400 categories.
It contains $\app$240k training videos and $\app$20k validation videos on which we report results.
Performance is measured by top-1 and top-5 accuracy.

\paragraph{Baseline Model and Training.}
We use a ResNet-50 (R50) SlowFast network~\cite{resnet,slowfast} with a 32-frame fast pathway, speed ratio $\alpha{=}$4, and channel ratio $\beta{=}$1$/$8 as our default model.
Input frames are sampled at a temporal stride of 2.

Our baseline training recipe follows Feichtenhofer~\etal~\cite{slowfast}.
We run synchronous SGD for 112k iterations on 128 GPUs with a mini-batch size of 4 clips per GPU ($\app$239 epochs) with initial learning rate of 0.8. (We perform single GPU experiments in \secref{casestudy}.) The learning rate is decreased by 10$\x$ at iterations 44k, 72k, and 92k.\footnote{We use the stepwise learning rate schedule rather than the cosine schedule used in Feichtenhofer~\etal~\cite{slowfast} because it is still more common. Results with a cosine schedule are available in the Appendix.}
We use a weight decay of 10$^{-4}$, momentum of 0.9, and a linear learning rate warm-up~\cite{goyal2017accurate} from 0.002 over 16k iterations.
Input clips are random 224$\x$224 spatial crops from clips that are randomly resized such that the shorter side  $\in [$256, 340$]$ pixels.

At test time, we sample 10 clips per video with uniform temporal spacing and combine the predictions with average pooling following~\cite{wang2018non,slowfast,korbar2019scsampler}.
We use 224$\x$224 center crop testing by default~\cite{korbar2019scsampler,wu2019long} and present results with other settings in the Appendix.

We select these training and inference procedures based on validation accuracy \emph{using the baseline training method}.
We adopt the exact same recipe for multigrid training experiments, aside from multigrid specific changes.
This choice may put multigrid training at a disadvantage, but it reflects the realistic scenario in which one wants to apply multigrid training to accelerate an already known training schedule without further tuning.

\paragraph{Evaluation.} Speedup factors are wall-clock GPU training time on P100 GPUs with CUDA 9.2 and cuDNN 7.6.3.
For fair comparison, the same hardware and software implementation is used for all methods.
We note that multigrid training exploits larger mini-batches, which increases data loading throughput requirements.
Training may become IO bound if the data loader is not optimized appropriately or if remote data access is used.
With sufficient local disk and an optimized data loader, training is typically not IO bound.

\paragraph{Multigrid Training Details.}
To sample data with spatial shape $h{\x}w$ that is smaller than $H{\x}W$, we change the default random short-side interval to $[$256$\frac{h}{H}$, 340$]$, noting that $w{=}h$ in our experiments. For the temporal dimension, we take $t$ ($t{<}T$) frames with random stride in $[$2, 2$\frac{T}{t}$$]$.

\subsection{Main Results}
We compare multigrid training to baseline training in \figref{exp:main}.
In addition to the default baseline, one could speed up training by using a smaller spatial-temporal shape with a larger mini-batch size and learning rate, so we also compare to this baseline variant.
For each method, we experiment with training schedules that range from 0.25$\x$ to 3$\x$ the number of baseline epochs ($\sim$239) to study the trade-off between training time and accuracy.
Overall, multigrid training always achieves a better trade-off than baseline training.
For example, multigrid training with both the long and short cycles can iterate through 1.5$\x$ more epochs than baseline method, while only requiring 1$/$3.4$\x$ the number of iterations, 1$/$4.5$\x$ training time, and achieving higher accuracy (75.6\% $\rightarrow$ 76.4\%).
The wall-clock speedup is greater than the iteration reduction factor, as a larger mini-batch with smaller space/time dimensions is more parallelism-friendly on modern GPUs.
Both the long and short cycles improve the trade-off and using both together performs the best.

In \figref{exp:main} we also observe that baseline training suffers a decline in accuracy when training for $\ge$1.5$\x$ epochs. With either long and/or short cycles, a decrease in accuracy is not observed for schedules up to 2.0$\x$ epochs, indicating that variable grids can help prevent overfitting.

In the following we use multigrid training with long and short cycles and 1.5$\x$ more epochs than the baseline as our default since it obtains a good trade-off.

\subsection{Ablation Experiments}\label{sec:exp:ablation}

\paragraph{Long Cycle Design.}
By default, we use $S{=}4$ long cycle shapes with a 1.5$\x$ epoch schedule.
In \tabref{ablation:long}, we explore using fewer shapes, where we take the last $S'{<}S$ shapes, for $S'{\in}\{1,2,3\}$. The short cycle is used in these experiments, and the $S'{=}1$ setting is equivalent to using the short cycle only. We run all variants for the same number of training iterations (to roughly preserve total training FLOPs), noting that methods which use fewer shapes will process fewer epochs due to having smaller mini-batches on average compared to the $S{=}4$ design.

We see that using each additional shape improves accuracy and saturates at $S{=}4$ (default).
The improvement in accuracy is possibly due to the more examples seen by the model given the same amount of iterations.
Compared with $S{=}1$ (\ie, short cycle only), our default choice improves the top-1 accuracy
by absolute 2.4\% (74.0\% $\rightarrow$ 76.4\%), while being slightly faster (4.0$\x$ $\rightarrow$ 4.5$\x$).
All results use the fine-tuning phase, which we find is beneficial to varying degrees in different settings.
With the default schedule, it leads to 0.4\% absolute gain (76.0\% $\rightarrow$ 76.4\%; not shown in table).

\paragraph{Short Cycle Design.}
Adding each input shape to the short cycle leads to a clear accuracy improvement, \tabref{ablation:short}.
Our default short cycle design (3-shape) improves over 1-shape (\ie, no short cycle / long-cycle only) by
absolute 1.9\% (74.5\% $\rightarrow$ 76.4\%) in top-1 accuracy.

\begin{table*}[t]
\vspace{-3mm}
\hfill
  \subfloat[\textbf{Long cycle design (with default short cycle)}\label{tab:ablation:long}]{%
\tablestyle{2.5pt}{1.12}\begin{tabular}{@{}llx{24}x{22}x{22}@{}}
  & long cycle design & speedup & top-1 & top-5\\
\shline
  Baseline & - & - & 75.6 & 91.9\\
  \hline
  \multirow{4}{*}{Multigrid}&1-shape (short cycle only) & 4.0$\x$ & 74.0 & 91.4\\
  &2-shape & 4.3$\x$ & 75.5 & 92.1\\
  &3-shape & 4.4$\x$ & 76.2 & 92.4\\
  &\textbf{4-shape (default)} & \textbf{4.5}$\x$ & \textbf{76.4} & \textbf{92.4}\\
\end{tabular}}\hfill
  \subfloat[\textbf{Short cycle design (with default long cycle)}\label{tab:ablation:short}]{%
\tablestyle{2.5pt}{1.12}\begin{tabular}{@{}llx{24}x{22}x{22}@{}}
  & short cycle design & speedup & top-1 & top-5\\
\shline
  Baseline & - & - & 75.6 & 91.9\\
  \hline
  &1-shape (long cycle only) & 4.2$\x$ & 74.5 & 91.6\\
  Multigrid&2-shape & 4.3$\x$ & 75.5 & 92.1\\
  &\textbf{3-shape (default)} & \textbf{4.5}$\x$ & \textbf{76.4} & \textbf{92.4}\\
  & & & \\
\end{tabular}}\hfill
\vspace{2mm}
\caption{\textbf{Ablation Study.} We perform ablations on Kinetics-400 using an R50-SlowFast network.
We analyze the impact of the long cycle (\tabref{ablation:long}) and short cycle (\tabref{ablation:short}) designs. All variants of multigrid training use the same number of training iterations as our default 1.5$\x$ epoch schedule; this roughly preserves the total training FLOPs.
We report wall-clock speedup relative to the baseline trained for 1.0$\x$ epochs.
}\label{tab:ablations}
\end{table*}
\begin{table*}[h!]
\subfloat[\textbf{Pre-training}\label{tab:ablation:pretrain}]{%
\tablestyle{2.5pt}{1.12}\begin{tabular}{@{}lx{34}x{26}x{26}x{26}@{}}
  & pre-train? & speedup & top-1 & top-5\\
\shline
Baseline & & - & {75.6} & {91.9}\\
 \textbf{Multigrid} &  & \textbf{4.5}$\x$ & \textbf{76.4} & \textbf{92.4}\\
 \hline
 Baseline & \checkmark& - & 75.4 & {91.9}\\
 \textbf{Multigrid} & \checkmark & \textbf{4.5}$\x$ & \textbf{76.0} & \textbf{92.4}\\
 \\
 \\
\end{tabular}}\hfill
\subfloat[\textbf{Temporal shape $T$}\label{tab:ablation:temporal}]{%
\tablestyle{2.5pt}{1.12}\begin{tabular}{@{}x{26}lx{26}x{26}x{26}@{}}
  $T$ &  & speedup & top-1 & top-5\\
\shline
 16 & Baseline& - & 74.8 & 91.4\\
 16 & \textbf{Multigrid} & \textbf{4.0$\x$} & \textbf{75.2} & \textbf{91.9}\\
 \hline
 32 & Baseline & - & 75.6 & 91.9\\
 32 & \textbf{Multigrid} & \textbf{4.5$\x$} & \textbf{76.4} & \textbf{92.4}\\
 \hline
 64 & Baseline & - & 75.9 & 92.1\\
 64 & \textbf{Multigrid} & \textbf{5.5$\x$} & \textbf{77.6} & \textbf{93.2}\\
\end{tabular}}\hfill
\subfloat[\textbf{Spatial shape $H{\x}W$}\label{tab:ablation:spatial}]{%
\tablestyle{2.5pt}{1.12}\begin{tabular}{@{}x{30}lx{26}x{26}x{26}@{}}
  $H{\x}W$ & & speedup & top-1 & top-5\\
\shline
 224 & Baseline & - & 75.6 & 91.9\\
 224 & \textbf{Multigrid} & \textbf{4.5$\x$} & \textbf{76.4} & \textbf{92.4}\\
 \hline
 320 & Baseline & - & 75.1 & 91.8\\
 320 & \textbf{Multigrid} & \textbf{6.5$\x$} & \textbf{76.8} & \textbf{92.8}\\
 \\
 \\
\end{tabular}}
\vspace{2mm}
\caption{\textbf{Generalization Analysis.}
We study how multigrid training generalizes to models both with and without ImageNet pre-training (\tabref{ablation:pretrain}) and
models of different temporal (\tabref{ablation:temporal}) and spatial (\tabref{ablation:spatial}) shapes.
All experiments use R50-SlowFast with results on Kinetics-400.
We use the default setting for multigrid training (1.5$\x$ more epochs, corresponding to 3.4$\x$ fewer iterations than baseline) in all settings.
We observe that the default choice brings consistent speedup and performance gain in all cases.
}%
\label{tab:generalization}
\end{table*}

\subsection{Generalization to Different Training Settings}\label{sec:exp:generalization}
Next we study how multigrid training generalizes to different training settings that are common in practice.

\paragraph{Pre-training.}
In our main results, we train models from random initialization.
We see in \tabref{ablation:pretrain} that with ImageNet~\cite{Russakovsky2015} pre-training, our multigrid method obtains a similar speedup and performance gain.
(We will present more results on ImageNet-pre-trained models in \secref{i3d}.)

\paragraph{Temporal Shape.}
Next we show generalization of multigrid training for models of different temporal shapes $T$.
We compare models that use 16-frame, 32-frame (default), and 64-frame input clips.\footnote{$\alpha{=}2$ in the 16-frame model to avoid a degenerated slow pathway.}
In all cases (\tabref{ablation:temporal}), multigrid training achieves a consistent accuracy gain and speedup.
The 64-frame model enjoys the largest performance gain (75.9\% $\rightarrow$ 77.6\%) and the best speedup (5.5$\x$).

\paragraph{Spatial Shape.}
We also demonstrate generalization of our method for models of different spatial shapes $H{\x}W$.
We increase the baseline shape from 224$\x$224 (default) to 320$\x$320 and study the impact.
Inference for the 320$\x$320 model is analogous to the 224$\x$224 case; we resize shorter side to 352 pixels and test on center 320$\x$320 crops.
In \tabref{ablation:spatial}, we see that multigrid training leads to an even larger performance gain (75.1\% $\rightarrow$ 76.8\%) and a more significant speedup (6.5$\x$) in the 320$\x$320 case.
Also note with the baseline method 320$\x$320 does not work better than 224$\x$224,
possibly due to overfitting, similar to what is reported in Tan~\etal~\cite{tan2019efficientnet}.
On the other hand, with multigrid training, spatial scaling brings improvement,
possibly due to the data augmentation brought by multigrid training.

\subsection{Generalization to Different Models}\label{sec:i3d}

So far we have focused on state-of-the-art SlowFast network~\cite{slowfast} for analysis.
We next demonstrate generalization of multigrid training to different networks by presenting results using a standard R50-I3D model~\cite{resnet,carreira2017quo} and its extension with non-local blocks (I3D-NL)~\cite{wang2018non}.

\paragraph{Implementation Details.}
Both models are ImageNet-pre-trained with 3D convolutions inflated from 2D convolutions following common practice~\cite{carreira2017quo,feichtenhofer2016spatiotemporal,wang2018non}.
Each input clip consists of 16 frames, sampled at a stride of 4.
I3D-NL additionally contains 5 (dot product) non-local blocks~\cite{wang2018non} in res$_3$ and res$_4$ stages.
The exact model specification is given in the Appendix.

The baseline recipe trains for 100k iterations using 128 GPUs, with a mini-batch size of 2 clips per GPU ($\app$106 epochs)
and a learning rate of 0.04, which is decreased by a factor of 10 at iteration 37.5k and 75k.
We do not use learning rate warm-up~\cite{goyal2017accurate} following prior work~\cite{wang2018non}.
Other training details are analogous to SlowFast training.
We note again that this training recipe is selected to be the best for the baseline training method and we apply multigrid training on top without further tuning.

\paragraph{Evaluation.}
We summarize the results in \tabref{exp:i3d}.
For both I3D and I3D-NL, multigrid training with the default schedule (1.5$\x$ epoch) obtains similar or better accuracy, while being up to 3.9$\x$ faster.
We also experiment with a shorter baseline schedule (`baseline $\frac{1}{3.3}$' in table), which trains for the same number of iterations as the multigrid training.
The shorter baseline schedule obtains a lower accuracy (3.7\% and 3.2\% absolute top-1 lower than multigrid).
We also see that I3D-NL has a lower speedup than I3D.
This is in part due to the less optimized NL operator than convolution, consuming a large portion of the training time.
We observe consistent improvements with larger backbone models (R101); see the Appendix.

\begin{table}
\tablestyle{7pt}{1.12}\begin{tabular}{@{}llx{25}x{23}x{23}@{}}
 model & & speedup & top-1 & top-5\\
\shline
 I3D& Baseline & - & {74.4} & {91.4}\\
 I3D & Baseline $\frac{1}{3.3}$ & 3.3$\x$ & 71.1 & 89.9\\
 I3D & \textbf{Multigrid} & \textbf{3.9}$\x$ & \textbf{74.8} & \textbf{91.7}\\
 \hline
 I3D-NL & Baseline & - & \textbf{75.5} & {92.1}\\
 I3D-NL & Baseline $\frac{1}{3.3}$ & \textbf{3.3$\x$} & 72.3 & 90.6\\
 I3D-NL & \textbf{Multigrid} & \textbf{3.3$\x$} & \textbf{75.5} & \textbf{92.4}\\
\end{tabular}
\vspace{2mm}
\caption{\textbf{Kinetics-400 accuracy with I3D and I3D-NL.}
While developed on SlowFast~\cite{slowfast},
multigrid training provides a consistent speedup and performance gain with I3D~\cite{carreira2017quo}
and I3D-NL~\cite{wang2018non}.
}\label{tab:exp:i3d}
\end{table}

\section{Case Study: 1-GPU Training on Kinetics}\label{sec:casestudy}
Our experiments thus far use a large number of GPUs (128) in parallel. However, a more common training recipe may use far fewer GPUs (\eg, 1 to 8) and given that one of our goals is to make video research more accessible by reducing computational requirements it is important to explore the application of our multigrid method in the few-GPU regime, without any tuning.

As a case study, we use a \emph{single GPU} to train an I3D model on Kinetics-400 using the quick training recipe from the public repository\footnote{\url{https://github.com/facebookresearch/video-nonlocal-net}} of Wang~\etal~\cite{wang2018non}. We apply multigrid training on top without further tuning.
This schedule trains for 1200k iterations (after adjusting with the linear scaling rule~\cite{goyal2017accurate}) on one GPU with 8 clips ($\app$40 epochs).
The learning rate is 0.00125, which is decreased by a factor of 10 at iteration 600k and 1000k.
Dropout and random scaling are disabled to accelerate convergence given the short schedule.
Each input clip consists of 8 frames, sampled at a stride of 8, when using the baseline optimizer.
Other training details are the same as the I3D experiments.

\tabref{exp:1gpu} shows that multigrid training generalizes well out-of-the-box to a few-GPU, short-schedule setting.
With multigrid training we are able to achieve 72.5\% (73.1\% with 30-crop testing~\cite{wang2018non}) top-1 accuracy \emph{in 2 days using only 1 GPU}, while the baseline method would need nearly 1 week. (When using a small model, we observe a smaller wall-clock speedup of $\app$3.3$\x$ compared to a larger model, which typically yields a $\app$4.5$\x$ speedup).
We hope the reduced training time with multigrid training will make video understanding research more accessible and economical.

\begin{table}
\tablestyle{7.5pt}{1.12}\begin{tabular}{@{}lx{63}x{22}x{22}@{}}
 & training time (days)& top-1 & top-5\\
\shline
 Baseline & 6.7 & 72.5 & 90.4\\
 \textbf{Multigrid} & \textbf{2.0} & {72.5} & {90.4}\\
\end{tabular}
\vspace{2mm}
\caption{\textbf{Case study: 1-GPU training on Kinetics-400.}
Multigrid training reduces the training time from nearly 1 week to 2 days on a single GPU\@.
We hope the reduced training time will make video understanding research more accessible and economical.
}\label{tab:exp:1gpu}
\end{table}

\section{Experiments on Something-Something V2}\label{sec:ssv2}

We next evaluate multigrid training on the Something-Something V2 dataset~\cite{ssv2},
which contains 169k training, and 25k validation videos.
Each video shows an interaction with everyday objects.
The task is classification with 174 action classes.
Performance is evaluated by top-1 and top-5 accuracy.
This task is known to require more `temporal modeling' to solve than Kinetics~\cite{xie2018rethinking}.

\paragraph{Implementation Details.}
We use an R50-SlowFast model~\cite{resnet,slowfast} with 64-frame fast pathway with speed ratio $\alpha{=}$4 and channel ratio $\beta{=}$1$/$8.
The model is pre-trained on Kinetics-400 following prior work~\cite{lin2018temporal}.
The baseline training recipe trains for 230k iterations on 8 GPUs, with a mini-batch size of 2 clips per GPU
and a learning rate of 0.03, which is decreased by a factor of 10 at iteration 150k and 190k.
Other training details are analogous to Kinetics experiments; see the Appendix for details.

\paragraph{Results.}
Similar to what we observe on Kinetics, multigrid training obtains a better trade-off than baseline training on Something-Something V2 (\tabref{ssv2}).
With the default 1.5$\x$-epoch training, multigrid training is 5.6$\x$ faster while obtaining a slightly higher accuracy. Multigrid training behaves consistently for the `spatial heavy' Kinetics dataset
and the `temporal heavy' Something-Something V2 dataset.

\begin{table}[t]
\tablestyle{7.5pt}{1.12}\begin{tabular}{@{}lx{35}x{25}x{25}@{}}
 & speedup & top-1 & top-5\\
\shline
 Baseline & - & 60.9\mypm{0.31} & 87.2\mypm{0.13}\\
 \hline
 Baseline $\frac{1}{5.2}$& 5.2$\x$ & 54.6\mypm{0.13} & 83.0\mypm{0.14}\\
 \textbf{Multigrid} 1.0$\x$& \textbf{8.3}$\x$ & 60.0\mypm{0.31} & 86.8\mypm{0.05}\\
 \hline
 Baseline $\frac{1}{3.4}$& 3.4$\x$ & 57.3\mypm{0.13} & 84.7\mypm{0.15}\\
 \textbf{Multigrid} 1.5$\x$ (default)& 5.6$\x$ & 61.2\mypm{0.18} & 87.4\mypm{0.12}\\
 \hline
 Baseline $\frac{1}{2.6}$& 2.6$\x$ & 58.7\mypm{0.06} & 85.8\mypm{0.15}\\
 \textbf{Multigrid} 2.0$\x$& 4.2$\x$ & \textbf{61.7\mypm{0.20}} & \textbf{87.8\mypm{0.12}}\\
\end{tabular}
\vspace{2mm}
\caption{\textbf{Results on Something-Something V2.}
Multigrid training achieves a better trade-off than baseline training.
Results are the mean and standard deviation over 5 runs.
\vspace{2.5mm}
}\label{tab:ssv2}
\end{table}

\section{Experiments on Charades}\label{sec:charades}
\begin{table}[t]
\tablestyle{14pt}{1.12}\begin{tabular}{@{}lx{35}x{35}@{}}
 & speedup & mAP (\%)\\
\shline
 Baseline & - & 38.0\mypm{0.18}\\
 \hline
 Baseline $\frac{1}{5.3}$& 5.3$\x$ & 27.5\mypm{0.15} \\
 \textbf{Multigrid} 1.0$\x$& \textbf{8.6}$\x$ & 36.8\mypm{0.31}\\
 \hline
 Baseline $\frac{1}{3.5}$& 3.5$\x$ & 31.5\mypm{0.26}\\
 \textbf{Multigrid} 1.5$\x$ (default)& {5.7}$\x$ & \textbf{38.2}\mypm{0.06}\\
 \hline
 Baseline $\frac{1}{2.6}$& 2.6$\x$ & 33.6\mypm{0.13}\\
 \textbf{Multigrid} 2.0$\x$& {4.3}$\x$ & 37.4\mypm{0.15}\\
\end{tabular}
\vspace{2mm}
  \caption{\textbf{Results on Charades.}
Multigrid training shows consistent speedups compared with the other datasets.
Results are the mean and standard deviation over 5 runs.
}\label{tab:charades}
\end{table}

We finally evaluate our method on the Charades dataset~\cite{charades}, which is relatively small, consisting of only 9,848 videos in 157 action classes.
The task is to predict all actions in a video.
Performance is measured by mAP.

\paragraph{Implementation Details.}
We use the same R50-SlowFast model~\cite{resnet,slowfast},
with the same Kinetics pre-training as the Something-Something experiments.
Training details are available in the Appendix.

\paragraph{Results.}
Overall we observe consistent results compared with Kinetics and Something-Something V2 (\tabref{charades}).
The default multigrid training is 5.7$\x$ faster, while achieving slightly better mAP\@.
Overall, we see that even for the smaller Charades dataset, with strong large-scale pre-training,
multigrid training is beneficial.

\section{Conclusion}
We propose a multigrid method for fast training of video models.
Our method varies the sampling grid and the mini-batch size during training,
and can process the same number of epochs using a small fraction of the computation of the baseline trainer.
With a single \mbox{\emph{out-of-the-box} setting}, it works on multiple datasets and models, and consistently brings a \mbox{$\app$3-6$\x$} speedup with comparable or higher accuracy.
It works across a spectrum of hardware settings from 128 GPU distributed training to \mbox{single GPU} training.
We hope the reduced training time will make video understanding research more accessible, scalable, and economical.

\newcount\cvprrulercount
\appendix
\section{Appendix}
\subsection{Supplementary Experiments}
\paragraph{R101-SlowFast Results.}
We demonstrate generalization of multigrid training to deeper backbones
by extending our default R50-SlowFast network to R101-SlowFast.
All other designs and training procedures remain unchanged.

\begin{table}[h]
\tablestyle{3pt}{1.12}\begin{tabular}{@{}llx{26}x{26}x{26}@{}}
  backbone & & speedup & top-1 & top-5\\
\shline
 R50 (default) & Baseline & - & 75.6 & 91.9\\
 R50 (default) & \textbf{Multigrid} & \textbf{4.5$\x$} & \textbf{76.4} & \textbf{92.4}\\
 \hline
 R101 & Baseline & - & 76.5 & 92.4\\
 R101 & \textbf{Multigrid} & \textbf{4.4$\x$} & \textbf{77.0} & \textbf{92.9}\\
\end{tabular}
\end{table}

As expected, R101-SlowFast outperforms R50-SlowFast and we observe a consistent speedup and accuracy gain over the baseline with multigrid training.

\paragraph{Long Cycle Design.}
By default we use multiple long cycles that are synchronized with the stepwise learning rate (LR) schedule (\ie, one long cycle period per LR stage).
We compare our default design (`multi-cycle') with an alternative that uses only a single long cycle period (`single-cycle') throughout all of training. Note that the single-cycle design does not use a fine-tuning phase as it is unclear how to incorporate it into this design.

\begin{table}[h]
\tablestyle{2.5pt}{1.12}\begin{tabular}{@{}llx{26}x{26}x{26}@{}}
  & long cycle design & speedup & top-1 & top-5\\
\shline
 Baseline & - & - & 75.6 & 91.9\\
 \hline
 \multirow{2}{*}{Multigrid} & single-cycle & 5.2$\x$ & 74.4 & 91.8\\
  & \textbf{multi-cycle (default)} & 4.5$\x$ & \textbf{76.4} & \textbf{92.4}\\
\end{tabular}
\end{table}

We observe that our default, multi-cycle design works better. In the multi-cycle design, the later shapes, which are closer to the final testing distribution, are used with each LR. We conjecture that exposing the model to these shapes with the larger (earlier) LRs is important for generalizing to the testing distribution. In contrast, the single-cycle design only uses the later shapes with relatively low LRs.

\paragraph{Cosine Learning Rate Schedule.}
We develop multigrid training assuming a stepwise LR schedule.
Next we experiment with a cosine LR schedule.
We experiment with both the multi-cycle and the single-cycle design for long cycles.
\emph{No further modifications} are applied to multigrid training.

\begin{table}[h]
\tablestyle{2.5pt}{1.12}\begin{tabular}{@{}llx{24}x{24}x{24}@{}}
 LR schedule & & speedup & top-1 & top-5\\
\shline
 \multirow{2}{*}{Stepwise (default)} & Baseline & - & 75.6 & 91.9\\
 & Multigrid & \textbf{4.5$\x$} & \textbf{76.4} & \textbf{92.4}\\
 \hline
 \multirow{2}{*}{Cosine} & Baseline & - & 75.8 & 92.0\\
 & Multigrid (single long cyc.) & \textbf{5.2}$\x$ & 75.4 & 92.1\\
 & Multigrid (multi long cyc.) & 4.2$\x$ & 75.3 & 92.1\\
\end{tabular}
\end{table}

We observe that multigrid training on a cosine schedule obtains a slightly lower accuracy than the default stepwise schedule.
The lower accuracy is possibly due to the relatively smaller learning rates used in larger (later) shapes as the LR is monotonically decreasing in a cosine schedule.
However, it still obtains a consistent speedup and a comparable accuracy to baseline, suggesting robustness of the multigrid strategy.
The two long-cycle designs obtain a similar accuracy.

\paragraph{Testing Settings}
Next we present results with additional test-time settings that are common in the literature.
Here we use the 64-frame R50-SlowFast due to its high accuracy. Our multigrid method trains this model 5.5$\x$ faster than the baseline.
\begin{table}[h!]
\tablestyle{3pt}{1.12}\begin{tabular}{@{\extracolsep{3pt}}lx{20}x{20}x{20}x{20}x{20}x{20}@{}}
 & \multicolumn{2}{c}{center 224$^{2}$} & \multicolumn{2}{c}{3-crop 224$^{2}$} & \multicolumn{2}{c}{3-crop 256$^{2}$}\\
 \cline{2-3}  \cline{4-5}  \cline{6-7}
 & top-1 & top-5 & top-1 & top-5 & top-1 & top-5\\
\shline
Baseline & 75.9 & 92.1 & 76.5 & 92.2 & 77.2 &  92.5\\
\textbf{Multigrid} & \textbf{77.6} & \textbf{93.2} & \textbf{78.1} & \textbf{93.5} & \textbf{78.1} & \textbf{93.4}
\end{tabular}
\end{table}

As expected, using 3-crop (left-center-right) testing improves accuracy for both baseline and multigrid training.

\subsection{Supplementary Implementation Details}
The I3D and I3D-NL architectures used in generalization analysis are shown below (assuming 16$\x$224$\x$224 inputs):

\newcommand{\blockb}[3]{\multirow{3}{*}{\(\left[\begin{array}{c}\text{1$\x$1$\x$1, #2}\\[-.1em] \text{1$\x$3$\x$3, #2}\\[-.1em] \text{1$\x$1$\x$1, #1}\end{array}\right]\)$\x$#3}}
\newcommand{\blockbt}[3]{\multirow{3}{*}{\(\left[\begin{array}{c}\text{3$\x$1$\x$1, #2}\\[-.1em] \text{1$\x$3$\x$3, #2}\\[-.1em] \text{1$\x$1$\x$1, #1}\end{array}\right]\)$\x$#3}}
\begin{table}[h]
\footnotesize
\tablestyle{9.2pt}{1.05}
\begin{tabular}{@{}llc@{}}
Layer & \multicolumn{1}{c}{Specification} & Output size \\
\shline
conv$_1$ & \multicolumn{1}{c}{1$\x$7$\x$7, 64, stride 1, 2, 2} & 16$\x$112$\x$112 \\
\hline
pool$_1$  & \multicolumn{1}{c}{1$\x$3$\x$3 max, stride 1, 2, 2\ \ \ } & 16$\x$56$\x$56 \\
\hline
\multirow{3}{*}{res$_2$} & \blockb{256\ \ }{64\ \ \ \ }{3} & \multirow{3}{*}{16$\x$56$\x$56} \\
  &  & \\
  &  & \\
\hline
\multirow{3}{*}{res$_3$} & \blockb{512\ \ }{128\ \ }{4} & \multirow{3}{*}{16$\x$28$\x$28} \\
  &  & \\
  &  & \\
\hline
\multirow{3}{*}{res$_4$} & \blockbt{1024}{256\ \ }{6} & \multirow{3}{*}{16$\x$14$\x$14}  \\
  &  & \\
  &  & \\
\hline
\multirow{3}{*}{res$_5$} & \blockbt{2048}{512\ \ }{3} & \multirow{3}{*}{16$\x$7$\x$7} \\
  &  & \\
  &  & \\
\end{tabular}
\end{table}

`I3D-NL' additionally uses non-local operators~\cite{wang2018non} after blocks 1 and 3 of res$_\textrm{3}$, and blocks 1, 3, and 5 of res$_\textrm{4}$.

\paragraph{Something-Something V2 Training.}
We use a linear warm-up~\cite{goyal2017accurate} for 2k iterations from 0.0001 and a weight decay of 10$^{-6}$.
As Something-Something V2 requires distinguishing between directions, we disable random flipping during training.
Following \cite{lin2018temporal}, we use segment-based input frame sampling, \ie, we split each video into segments, and from each of them, sample one frame to form a clip.

\paragraph{Charades Training.}
The baseline method trains for 28k iterations with a learning rate of 0.0375,
which is decreased by a factor of 10 at iteration 20k and 24k.

\paragraph{Acknowledgments.}
We would like to thank Haoqi Fan for helping with the code release
and Ishan Nigam, Santhosh K. Ramakrishnan, and Xingyi Zhou for helpful comments on an earlier draft.
This material is partially supported by the National Science Foundation under Grant No.\ IIS-1845485
and the Facebook Fellowship to C-Y Wu.

{\small
\bibliographystyle{ieee_fullname}
\bibliography{multigrid}
}

\end{document}